\documentclass[sigconf]{aamas} 

\usepackage{balance} 
\usepackage{booktabs}
\usepackage{tabularx}
\usepackage{xcolor}
\usepackage[framemethod=TikZ]{mdframed}

\definecolor{bluesky}{RGB}{100, 170, 255}

\mdfsetup{skipabove=\topskip,skipbelow=\topskip}
\newenvironment{informaldef}[1][]{%
\ifstrempty{#1}%
{
\mdfsetup{%
innertopmargin=5pt
}
}%
{\mdfsetup{%
innertopmargin=2pt,%
frametitle={%
\tikz[baseline=(current bounding box.east),outer sep=0pt]
\node[anchor=east,rectangle,fill=bluesky!30!white]
{\strut #1};}}%
}%
\mdfsetup{linecolor=bluesky,%
linewidth=2pt,topline=true,
frametitleaboveskip=\dimexpr-\ht\strutbox\relax,}
\begin{mdframed}[]\relax%
}{\end{mdframed}}

\makeatletter
\gdef\@copyrightpermission{
  \begin{minipage}{0.2\columnwidth}
   \href{https://creativecommons.org/licenses/by/4.0/}{\includegraphics[width=0.90\textwidth]{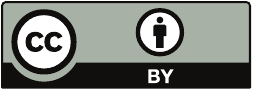}}
  \end{minipage}\hfill
  \begin{minipage}{0.8\columnwidth}
   \href{https://creativecommons.org/licenses/by/4.0/}{This work is licensed under a Creative Commons Attribution International 4.0 License.}
  \end{minipage}
  \vspace{5pt}
}
\makeatother

\setcopyright{ifaamas}
\acmConference[AAMAS '26]{Proc.\@ of the 25th International Conference
on Autonomous Agents and Multiagent Systems (AAMAS 2026)}{May 25 -- 29, 2026}
{Paphos, Cyprus}{C.~Amato, L.~Dennis, V.~Mascardi, J.~Thangarajah (eds.)}
\copyrightyear{2026}
\acmYear{2026}
\acmDOI{}
\acmPrice{}
\acmISBN{}

\acmSubmissionID{bst091}

\subtitle{Blue Sky Ideas Track}

\title[Foundation World Models]{Foundation World Models for Agents that Learn, Verify, and Adapt Reliably Beyond Static Environments
}

\author{Florent Delgrange}
\affiliation{
  \institution{AI Lab, Vrije Universiteit Brussel \& Flanders Make}
  \city{Brussels}
  \country{Belgium}
}

\begin{abstract}
The next generation of autonomous agents must not only learn efficiently but also act reliably and adapt their behavior in open worlds. Standard approaches typically assume fixed tasks and environments with little or no novelty, which limits world models' ability to support agents that must evolve their policies as conditions change. This paper outlines a vision for \textbf{foundation world models}: persistent, compositional representations that unify reinforcement learning, reactive/program synthesis, and abstraction mechanisms.
We propose an agenda built around four components:
(i) learnable reward models from specifications to support optimization with clear objectives;
(ii) adaptive formal verification integrated throughout learning;
(iii) online abstraction calibration to quantify the reliability of the model’s predictions; and
(iv) test-time synthesis and world-model generation guided by verifiers.
Together, these components enable agents to synthesize verifiable programs, derive new policies from a small number of interactions, and maintain correctness while adapting to novelty. The resulting framework positions foundation world models as a substrate for learning, reasoning, and adaptation, laying the groundwork for agents that not only act well but can explain and justify the behavior they adopt.
\end{abstract}

\keywords{Reinforcement Learning, Reactive Synthesis, Formal Verification, Foundation Models, World Models, Abstraction, Adaptive Agents}

\newcommand{\BibTeX}{\rm B\kern-.05em{\sc i\kern-.025em b}\kern-.08em\TeX}

\begin{document}


\pagestyle{fancy}
\fancyhead{}


\maketitle 


\section{Introduction}
In the last decade, \emph{reinforcement learning} (RL \cite{sutton_reinforcement_2018}) has achieved remarkable empirical progress, from surpassing human performance in high-dimensional games \cite{mnih_human-level_2015,hafner_mastering_2021,hafner_training_2025} to mastering complex real-world control tasks such as energy management, robotics, epidemics mitigation \cite{xiao_motion_2022,degrave_magnetic_2022,haarnoja_learning_2024,reymond_exploring_2024}.
Beyond control and decision-making, RL has also become central to the post-training of large language models (LLMs).
Techniques such as RL from human feedback, verifiable rewards, and other variants \cite{christiano_deep_2017,ouyang_training_2022,lambert_tulu_2024,lee_rlaif_2024}
have turned RL into a general framework for aligning agents with qualitative goals such as coherence, safety, or reasoning depth.
These advances, however, have largely relied on reward maximization and massive data collection, with little understanding of the structure underlying the learned agent's behaviors.
Aligning agents' intended behaviors with designed or learned reward models remains deeply challenging: it is prone to reward hacking, fragile to misspecification, and often demands extensive engineering.
In addition, modern approaches couple RL with deep neural networks to scale to large domains.
As a result, learning guarantees no longer hold, policies may drift from the designer’s intent, and fail to respect safety constraints.
Despite numerous efforts to regularize or constrain learning \cite{junges_safety-constrained_2016,alshiekh_safe_2018,jansen_safe_2020,yang_safe_2023,zikelic_learning_2023,zikelic_compositional_2023,debot_neurosymbolic_2025,delgrange_composing_2025,badings_policy_2025,henzinger_supermartingale_2025}, the ability to guarantee correctness remains fundamentally elusive. The question is no longer whether agents learn efficiently, but whether they learn reliably.

In contrast, \emph{reactive synthesis} \cite{pnueli_synthesis_1989,baier_model_2018} offers a complementary paradigm rooted in formal methods. Given a model of the environment (e.g., a \emph{Markov process} \cite{puterman_markov_1994}) and a logical specification of the desired behavior, synthesis algorithms construct control policies that are correct by design. These methods guarantee that the induced agent's behaviors satisfy the specification, providing the transparency and guarantees missing from most learning-based systems. Yet, their scope remains limited: the need for explicit, finite models of the environment and the cost of exhaustively exploring large state spaces to construct guaranteed policies render them impractical for the open-ended, general, uncertain domains where learning excels. The two fields thus stand as mirror images: one efficient, scalable but ungrounded, the other reliable but rigid.

\emph{Bridging these paradigms} is a defining challenge for the future of the (Multi-)Agent Systems community: in many multi-agent domains, ranging from marketplaces to autonomous mobility, agents must adapt to uncertainty while still satisfying global coordination and safety constraints. Purely reward-driven multi-agent RL often yields unstable or undesired behaviors, while purely synthesized policies rely on fixed models and rigid interaction assumptions \cite{Chopra2018HandbookON,DBLP:journals/aamas/ZhuDW24}. Recent integrations of multi-agent learning with temporal-logic objectives \cite{DBLP:conf/atal/HammondA0W21,DBLP:conf/ecai/HahnPSS0W24} highlight the need for methods that ensure adaptation and correctness. A unified perspective would support flexible learning of interaction patterns while maintaining verifiable guarantees over joint behavior in open multi-agent systems.

\begin{figure*}[t]
    \centering
    \includegraphics[width=.725\linewidth]{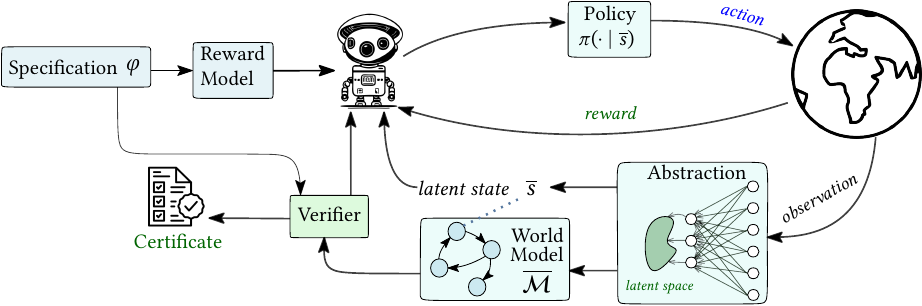}
    \vspace{-.5em}
    \caption{
    Given a \emph{specification} $\varphi$ formalizing the intended agent's behavior,\protect\footnotemark~a \emph{reward model} is automatically converted/generated from $\varphi$. Simultaneously with optimizing the return, the agent learns a representation of the observations through the state space of a \emph{verifiable} world model. Consequently, the policy is learned directly on this representation. To provide the guarantees, the world model is processed through a verifier (e.g., a model-checker). The verifier's output \emph{guides} the agent or \emph{corrects} the policy as needed. At any time, the verifier can return a certificate on both specification satisfaction and the abstraction quality of the world model.
    Note that we do not assume to have access to the explicit environment dynamics (simulating it is sufficient).
}
    \label{fig:framework}
\end{figure*}

Beyond theoretical convergence, current AI trends make guaranteed policy synthesis urgent. The rise of \textbf{foundation models} and agentic AI frameworks has renewed interest in systems that reason over goals, tools, and environments expressed in natural language.
First, adaptive and unsupervised RL highlight the need for representations that persist across tasks and agent populations, rather than being retrained from scratch \cite{touati_learning_2021,DBLP:conf/icml/AlegreB022,park_metra_2024,DBLP:conf/iclr/TirinzoniTFGKXL25,DBLP:conf/icml/AgarwalS0025,sikchi_fast_2025}.
In this context, the ``train-then-verify'' paradigm \cite{bacci_probabilistic_2020,delgrange_distillation_2022,bacci_verified_2022,delgrange_wasserstein_2023,delgrange_activating_2024}, in which a policy is first learned and later verified for compliance, remains conceptually unsatisfactory and computationally brittle in \emph{dynamic environments}, \emph{evolving goals}, and \emph{distributed settings} where \textbf{agents must continuously adapt while preserving constraints}.
Second, LLMs already exhibit rudimentary forms of reactive behavior: they interpret instructions, generate sequences of actions, and adapt decisions from interaction feedback \cite{DBLP:conf/iclr/YaoZYDSN023}.
Yet, they lack the formal backbone needed for correctness and long-term reliability. Integrating LLM-based reasoning with formal verification --- by treating language as a medium for generating, refining, and testing specifications --- offers a route toward interpretable and verifiable intelligence. For instance, LLMs can act as specification refiners, translating high-level objectives into temporal-logic constraints or decomposing tasks into verifiable subtasks (e.g., \cite{castanyer_arm-fm_2025}).

\footnotetext{$\varphi$ can be provided by the user or be \emph{generated} from experiences by an LLM.}

\paragraph{The vision: learning verifiable world models.}
We argue that the next generation of intelligent agents should \textbf{\emph{learn}} \textbf{verifiable world models}: internal representations that are not only \textbf{predictive} but also \textbf{formally analyzable}. Learning and verification become intertwined: as agents interact with their environment, they refine a predictive model while building a symbolic abstraction on which reasoning and verification can operate. This joint structure enables efficient planning, online guarantees, and the ability to \textbf{detect, quantify, and correct} deviations from intended specifications. In this sense, it bridges empirical learning and formal reasoning and points toward reliable, adaptive, and self-improving (multi-)agent systems.

{This vision} builds upon emerging evidence that such integration is feasible. Recent research demonstrates that world models can be learned with measurable abstraction error \cite{delgrange_distillation_2022,delgrange_wasserstein_2023}, that \textbf{deep} model-based RL can exploit learned, local dynamics to achieve guaranteed safe policy improvement \cite{delgrange_deep_2025}, and that learned representations can be lifted to the symbolic level for compositional synthesis \cite{zikelic_compositional_2023,delgrange_composing_2025,piriyakulkij_poe-world_2025,castanyer_arm-fm_2025}. These developments suggest that RL and reactive synthesis need not be antagonistic paradigms but can form the two halves of a coherent framework for reliable autonomy.

\section{Learning Verifiable World Models}
Achieving reliability and adaptability in autonomous agents demands that learning, verification, and reasoning become parts of a single closed loop.
This direction is promising and aligns with recent agendas on safe AI \cite{dalrymple_towards_2024} and with industrial interest in runtime verification for machine learning systems \cite{delseny_white_2021}.
We envision agents that continuously \textbf{learn, verify, and refine} their own \textbf{internal models of the world}. These self-verifying agents will not rely on externally imposed models or post-hoc safety checks; instead, they will build world models that are formally analyzable as they learn them, preserving guarantees of correctness and safety throughout their evolution. \emph{The general framework is depicted in Fig.~\ref{fig:framework}.}

At the core of this vision lies the concept of \emph{RL-Synthesizers}: agents that treat learning itself as a synthesis problem.
Each interaction cycle with the environment will jointly improve a predictive model, strengthen a symbolic abstraction of that model, and test compliance with a formal specification. The boundary between learning to act and reasoning about action will vanish: logical reasoning becomes part of the learning process itself, guiding policy updates rather than evaluating them afterward.

As a concrete example, consider a package-delivery agent in a dynamic warehouse: it must deliver a package while avoiding collisions with workers and robots. In the following, we will use this thread to illustrate the components illustrated in Fig.~\ref{fig:framework}.

\paragraph{Learning from Formalized Rewards}
Reward design remains a persistent source of brittleness in RL. Translating human intent into scalar feedback often requires extensive tuning and rarely ensures that the resulting policy reflects the intended objective. A promising path is to replace manual shaping with formally specified \emph{reward models}: functions derived from logical or programmatic task descriptions that retain semantic clarity and verifiability.

Temporal-logic–based translations, such as those introduced in \cite{camacho_ltl_2019} and other approaches incorporating LTL reasoning \cite{pnueli_temporal_1977} into RL \cite{toro_icarte_teaching_2018,bozkurt_control_2020,alur_framework_2022,shao_sample_2023}, already point in this direction.
That way, in the package delivery thread, a property such as ``eventually deliver while always avoiding collisions'' can be translated into a reward model, aligning optimization and specification satisfaction.
Yet most approaches emphasize expressiveness over statistical learnability: in general, learning with traditional LTL objective is intractable \cite{bazille_global_2020,yang_tractability_2022}.
Learnable alternatives leverage reward machines but lead to extremely large structures \cite{alur_policy_2023}, impeding formal verification.
Recent studies \cite{delgrange_activating_2024} have shown that certain logics, for example, the discounted logic of De Alfaro et al. \cite{alfaro_discounting_2003}, retain much of LTL’s expressiveness while being computably efficient and \emph{PAC-learnable}. Despite these advantages, such learnable fragments have been largely overlooked in contemporary RL research. Revisiting them is timely: when coupled with world-model abstraction, they enable differentiable and compact reward formulations whose optimization directly aligns with satisfaction of the underlying specification.

Together, these developments suggest that future agents should treat reward generation itself as a formal synthesis problem: combining learnable logics, programmatic representations, and language-based refinement to ensure that optimization and verification operate on the same semantic ground.

\paragraph{Verification During Learning}
Recent advances have increasingly blurred the boundary between deep learning and verification.
Techniques such as \emph{probabilistic shielding} \cite{alshiekh_safe_2018,jansen_safe_2020,yang_safe_2023,debot_neurosymbolic_2025,konighofer_shields_2025} and the learning of \emph{neural certificates} \cite{zikelic_learning_2023,zikelic_compositional_2023,giacobbe_neural_2024,mandal_formally_2024,yu_neural_2025,henzinger_supermartingale_2025,badings_policy_2025} already embed formal reasoning into the training process. They aim to ensure safety or specification satisfaction as the agent learns, rather than verifying a static policy post hoc. Yet these methods still rest on strong assumptions: they either presuppose access to an accurate, explicit model of the safety aspects of the environment, require the full dynamics to be known and differentiable, or do not scale to large domains. In open-ended settings, such assumptions collapse, limiting applicability precisely where guarantees are most needed.

A promising direction is to treat verification not as a one-time property of a model but as a continuous process of \textbf{calibration} between the learned world model and the real environment. As the agent updates its internal model, it simultaneously evaluates how the satisfaction of the specification evolves under model uncertainty. This allows the agent to \textbf{track and quantify} the confidence in its own guarantees, yielding a form of \textbf{runtime formal reasoning}. When the satisfaction margin falls below a safe threshold, the verifier can steer exploration toward critical regions, trigger additional data collection, or refine the world model, effectively transforming guarantees into \textbf{adaptive control signals}.
In the delivery example, if traffic patterns shift and the safety margin drops, verifier feedback can reject risky updates and redirect exploration.

A concrete route to verification during learning may build on the theory of \emph{Safe Policy Improvement} (SPI \cite{thomas_high_2015-1,ghavamzadeh_safe_2016,laroche_safe_2019, simao_safe_2020,castellini_scalable_2023,wienhoft_more_2023}), which guarantees that new policies do not degrade performance relative to a baseline. Intuitively, SPI maintains a set of plausible world models consistent with the collected data and accepts a policy update only when improvement can be certified under \emph{all} such models. This makes SPI a canonical instance of learning and formal reasoning co-evolving. While effective in offline and tabular settings, these methods require extensive data collection and do not scale to general spaces. Extending this to online, more general settings, deep RL agents could maintain such safety guarantees as their models evolve.
While this extension is non-trivial, recent works enable such a deep analogue of SPI theorems \cite{delgrange_deep_2025}, but do not offer practical algorithms explicitly leveraging the world model to act; doing so would allow the agent to reason beyond the distribution of experiences on which the model was trained, which is a clear direction toward broader adaptability.

Such mechanisms point toward self-calibrating verification, where reasoning adapts jointly with learning. The verifier is no longer an external auditor but a co-evolving module that constrains policy updates according to the learned model's reliability. This turns verification into an active feedback mechanism: a continuous dialogue between learning, uncertainty, and formal correctness.

\paragraph{Abstraction and World-Model Calibration}
At the foundation of any verifiable learning process lies the question of representation: \emph{how to model the environment compactly without losing the behavioral properties that matter for reasoning}. For agents that learn and verify, abstraction is not a preprocessing step but a continuously maintained contract between the learned world model and the true environment. The reliability of any verification or synthesis component ultimately depends on how well this contract is calibrated.

Traditional abstraction mechanisms, such as \emph{bisimulation metrics} \cite{desharnais_metrics_2004}, quantify \emph{behavioral equivalence} by bounding discrepancies between transition probabilities and rewards. In modern RL, these metrics have been extended to learned latent representations \cite{gelada_deepmdp_2019,castro_mico_2021,van_der_pol_plannable_2020,zhang_learning_2021,avalos_wasserstein_2024}, enabling approximate verification of neural policies and symbolic reasoning over compressed dynamics \cite{delgrange_distillation_2022,delgrange_wasserstein_2023}. Yet current estimators of abstraction error are conservative and often static, assuming fixed data distributions or fully explored state spaces. As policies evolve, such assumptions no longer hold: the abstraction must remain mathematically valid while drifting away from the parts of the environment the agent actually visits.
For example, after training on the main corridor, the RL agent may identify a rarely visited shortcut as a high-reward choice. The world model predicts it as safe, but unseen forklift traffic can violate collision avoidance unless the verifier marks that route as uncertified.

A crucial step forward is to make abstraction certificates \textbf{adaptive}: updated online as the agent collects data and revises its world model. This involves estimating local abstraction error, uncertainty, or divergence and feeding this information back to the verifier and planner. The resulting coverage-calibrated world models would provide a principled basis for determining when model-based reasoning can be trusted. Within this framework, abstraction bounds act as dynamic confidence radii that govern rollout budgets, planning horizons, or the activation of verification routines.

\section{Foundation World Models}
Adaptive abstraction opens the door to a broader goal: \emph{building persistent, general-purpose world models for open environments}:
\vspace{-.5em}
\begin{informaldef}[]
We see \textbf{\color{bluesky!80!blue}foundation world models} as \textbf{persistent}, \textbf{structured} representations of open environments that unify \textbf{dynamics}, \textbf{abstractions}, and \textbf{semantic knowledge}.
They act as \textbf{reusable} substrates that can be queried \textbf{\emph{zero-}} or \textbf{\emph{few-shot}} to \textbf{synthesize or adapt policies}, \textbf{verify} specifications, and \textbf{guide behavior} in \textbf{unseen~situations}.
\end{informaldef}
\vspace{-1em}
Such models would capture not only predictive dynamics but also the logical and relational structure of the environment: how local behaviors compose, what invariants hold, and under which abstractions guarantees remain valid. The objective is to move from isolated models learned per task toward persistent, verifiable priors that generalize across tasks and domains.

A foundation world model differs from conventional predictive models in three ways.
First, \emph{it integrates abstraction and calibration} as first-class components: each learned representation carries an explicit reliability measure, linking model accuracy to formal guarantees.
Second, \emph{it encodes compositional structure}, allowing previously learned fragments (such as verified local controllers or programmatic dynamics) to be assembled into novel behaviors.
Third, \emph{it supports semantic querying}: by conditioning on formal or linguistic task descriptions, the model can generate consistent world abstractions or policy priors for unseen objectives.

Recent research in \emph{unsupervised RL} \cite{touati_learning_2021,park_metra_2024,sikchi_fast_2025} and \emph{compositional synthesis} \cite{zikelic_compositional_2023,delgrange_composing_2025,piriyakulkij_poe-world_2025,castanyer_arm-fm_2025} provide evidence that these properties are attainable. Metric-aware representation learning suggests that temporal distance and behavioral similarity can guide the formation of transferable abstractions; modular or programmatic world-model frameworks show that local dynamics can be encoded as reusable components; and hierarchical control architectures demonstrate how verified local policies can be synthesized into globally correct strategies. Together, these advances draw a path toward foundation world models that unify generalization, structure, and verifiability.

Such models underpin the integration of learning and verification. They provide an analyzable substrate on which logical specifications can be generated, verified, and composed, enabling agents to acquire new competencies without retraining by reasoning over the verifiable components already contained in their world model.

\paragraph{LLMs as Specification Refiners}
Even the most robust foundation world model cannot anticipate every new configuration of an open world. When confronted with novel dynamics or goals, agents must rapidly form structured hypotheses about how the world behaves and how their objectives should be reformulated \cite{chollet_measure_2019,ying_assessing_2025,chollet_arc-agi-2_2025}. Both classical RL and LLM-based agents struggle in this setting: fast \textbf{adaptation} often requires retraining, and newly encountered constraints must be rediscovered from scratch. What is missing is a mechanism for learning world models and specifications \textbf{at test time}, grounded in formal reasoning and prior knowledge.

We propose an interactive refinement loop in which an LLM collaborates with a verifier to construct and validate new world models on the fly. This loop enables agents to synthesize verifiable programs and derive new policies after only a few interactions in an unseen environment.
\begin{enumerate}
    \item \emph{Specification and decomposition.} When the agent encounters unseen environment's regions, the LLM proposes candidate task specifications or decompositions (e.g., temporal-logic formulas or automata) derived from human instructions or from the agent’s initial exploratory trajectories.
    \item \emph{Program generation.} The observations then serve as a basis for the LLM to generate \emph{verifiable} programs in a formal modeling language (e.g., \textsc{Prism} \cite{kwiatkowska_quantitative_2005}) that can be processed by probabilistic model checkers (e.g., \textsc{Storm} \cite{hensel_probabilistic_2022}). \textbf{These programs act as world models}, capturing hypothesized causal or temporal relationships observed in the new environment.
	\item \emph{Formal verification.} The verifier checks the validity of the generated programs against its specifications and optionally reports counterexamples or structural inconsistencies.
	\item \emph{Revision and synthesis.} The LLM integrates this feedback to refine the program or propose new sub-tasks. The world model-programs and subtasks can then be used for planning, RL, or reactive synthesis to derive ``low-level'' policies or exploratory behaviors in the newly encountered domain.
    \item \emph{Repeat.} Execute a low-level policy for a specific sub-task, collect new experiences, and repeat from step (1).
\end{enumerate}

This process mirrors human reasoning: the agent alternates between hypothesis generation, verification, and empirical testing. Successive refinements would yield world models that are both compact and verifiable, enabling fast adaptation without retraining.
In the package-delivery example, a blocked corridor invalidates part of the original specification; the LLM thus refines that specification accordinlgy, regenerates the corresponding program, and submits it for formal verification. Then, the resulting program can be used as a world model to guide the policy to avoid the blocked corridor.

Importantly, the verifier's feedback can also serve as a \emph{formally verified reward signal}.
Thus, this reward may be used to \emph{reinforce the LLM’s capacity} to produce consistent, valid, and executable programs in RL post-training.
This closes the learning loop: successful verifications provide positive reinforcement, while counterexamples penalize inconsistent hypotheses.

This mechanism enhances adaptability by letting agents synthesize new abstractions and policies in situ, guided by logical structure and formal feedback.
Yet, this interactive loop remains challenging: generated programs may be inconsistent, verification can bottleneck repeated refinement, and abstraction or specification drift can amplify errors; robust deployment therefore requires stable, efficient inconsistency detection and feedback integration.
\section{Conclusion}

The trajectory of AI agents has long oscillated between empirical efficiency and formal correctness. RL yields increasingly competent agents, yet their behavior can be difficult to trust and interpret, whereas formal methods offer guarantees but depend on fixed models. Our vision reconciles these traditions by shifting what agents should learn: not only a policy, but a verifiable understanding of the world.

Our agenda rests on four components: formalized rewards tie optimization to intended task semantics; verification during learning turns correctness into an adaptive control signal; abstraction and world-model calibration quantify when model-based reasoning can be trusted; and LLM-based refinement supports test-time synthesis of specifications and programs in open environments.

Together, these components define foundation world models: persistent, compositional, and verifiable representations that unify perception, reasoning, and formal analysis. For the (Multi-)Agent Systems community, this shifts the goal from optimizing behavior in fixed domains to designing architectures that construct and certify their own domains of competence.

\section*{Acknowledgments}
This work was realized under F.~Delgrange's VUB OZR mandate (VUB-OZR4417) and was supported by the “DESCARTES” iBOF project.
We are grateful to Ann Nowé, Guillermo A. Pérez, Willem Röpke, and Pieter Libin for insightful discussions and thoughtful feedback that helped shape this vision and research agenda.

\bibliographystyle{ACM-Reference-Format} 
\balance
\bibliography{bib}


\end{document}